\documentclass[twoside,11pt]{article}

\usepackage{jmlr2e}
\usepackage[T1]{fontenc}
\usepackage{mathptmx}
\usepackage{graphicx}
\usepackage{amssymb}
\usepackage{amsmath}
\usepackage{booktabs}
\usepackage{color}
\usepackage{url}

\DeclareMathAlphabet{\mathcal}{OMS}{cmsy}{m}{n}

\newif\ifworkingpaper
\workingpaperfalse    
\ifworkingpaper
 \makeatletter
 \def\jmlrheading#1#2#3#4#5#6{\def\ps@jmlrtps{\let\@mkboth\@gobbletwo%
     \def\@oddhead{\scriptsize \@date \hfill}%
     \def\@oddfoot{\scriptsize \copyright #2 #6. \hfill}%
     \def\@evenhead{}\def\@evenfoot{}}%
   \thispagestyle{jmlrtps}}
 \def\@starteditor{\def\@editor{}}
 \def\@endeditor{}
 \makeatother
\fi

\graphicspath{{figures/}{}}

\def\eat#1{}
\def\eg.{\mbox{e.}\mbox{g.}}
\def\ie.{\mbox{i.}\mbox{e.}}

\def\Span{{\mathrm{Span}}}
\def\t{^\top}
\def\of#1{^{(#1)}}

\def\argmin{\mathop{\rm arg\,min}}
\def\algA{{\sf A}}
\def\algB{{\sf B}}
\def\calS{{\cal S}}
\def\calF{{\cal F}}
\def\calG{{\cal G}}
\def\calSo{{\calS^{\mu,L}}(\Omega)}
\def\calFo{{\calF^{\mu,L}_n}(\Omega)}
\def\calSz{{\calS^{\mu,L}}(\ell_2)}
\def\calFz{{\calF^{\mu,L}_n}(\ell_2)}
\def\calSzg{{\calS^{\mu,L}_\gamma}(\ell_2)}
\def\bbbR{{\mathbb{R}}}

\def\pprime{{\prime\prime}}

\def\lambdamin{{\lambda_{\min}}}
\def\lambdamax{{\lambda_{\max}}}
\def\Sigmahat{{\hat{\Sigma}}}
\def\kappaf{\kappa_{f}}
\def\kappai{\kappa_{i}}
\def\inBrace#1{{\left\{{#1}\right\}}}
\def\inBrack#1{{\left[{#1}\right]}}
\def\inPar#1{{\left({#1}\right)}}

\def\of#1{{\scriptstyle[#1]}}
\newcommand{\inprod}[2]{\ensuremath{\left\langle #1, #2
    \right\rangle}}
\newcommand{\order}{\ensuremath{\mathcal{O}}}
\newcommand{\otil}{\ensuremath{\tilde{\order}}}
\newcommand{\E}{\ensuremath{\mathbb{E}}}
\newcommand{\domain}{\ensuremath{\mathcal{X}}}

\newenvironment{proof*}{\par\noindent{\bf Proof\ }}{\hfill\BlackBox\par\smallskip}

\def\TheLongTitle{A Lower Bound for the Optimization of Finite Sums}
\def\TheShortTitle{\TheLongTitle}
\def\TheAuthors{Alekh Agarwal and L\'eon Bottou}

\jmlrheading{}{}{}{}{}{\TheAuthors}
\ShortHeadings{\TheShortTitle}{Agarwal and Bottou}
\firstpageno{1}

\begin{document}

\title{\TheLongTitle}

\author{\name Alekh Agarwal
  \email alekha@microsoft.com\\
  \addr Microsoft Research, New York, NY.\\
  \name L\'eon Bottou 
  \email leon@bottou.org\\
  \addr Facebook AI Research, New York, NY.\\}

\editor{n/a}

\maketitle
\raggedbottom
\sloppy


\bigskip
\begin{center}
  \textcolor{red}{\relax
    \begin{tabular}{|c|}
      \hline
      ~\parbox{0.9\linewidth}{\hfil\large\bf Erratum
      }~\\[1.5ex]
      ~\parbox{0.9\linewidth}{\relax
        We call the attention of the reader that the lower bound
        established in this paper is established the case of a
        determnistic optimization algorithm only. This limitation,
        applies to both theorems~1 and~2. It arises because the
        resisting oracle construction works only for deterministic
        algorithm. This is an important because the best known
        algorithms discussed in section~2 are randomized algorithms
        (section~3 remains entirely valid).  We believe that a
        comparable lower bound holds for randomized optimzation
        algorithms and are working on a substantially different (and
        substantially more complex proof) for the randomized case.  We
        shall update this document as soon as we are entirely satisfied
        with the new proof.\\[0.3ex] }~\\ \hline
  \end{tabular}}
\end{center}
\bigskip


\begin{abstract}
  This paper presents a lower bound for optimizing a finite sum of $n$
functions, where each function is $L$-smooth and the sum is
$\mu$-strongly convex. We show that no algorithm can reach an error
$\epsilon$ in minimizing all functions from this class in fewer than
$\Omega(n + \sqrt{n(\kappa-1)}\log(1/\epsilon))$ iterations, where
$\kappa=L/\mu$ is a surrogate condition number. We then compare this
lower bound to upper bounds for recently developed methods
specializing to this setting. When the functions involved in this sum
are not arbitrary, but based on i.i.d. random data, then we further
contrast these complexity results with those for optimal first-order
methods to directly optimize the sum. The conclusion we draw is that a
lot of caution is necessary for an accurate comparison, and identify
machine learning scenarios where the new methods help
computationally.
\end{abstract}


\section{Introduction}

Many machine learning setups lead
to the minimization a convex function of the form
\begin{equation}
    x^*_f = \argmin_{x\in\domain} f(x), ~\mbox{with}~
    f(x) = \frac{\mu}{2} \|x\|^2 + \frac{1}{n}\sum_{i=1}^{n}
    g_i(x), 
    \label{eqn:f-defn}
\end{equation}
where $\domain$ is a convex, compact set. When the functions $g_i$ are
also convex, then the overall optimization problem is convex, and can
in principle be solved using any off-the-shelf convex minimization
procedure. In the machine learning literature, two primary techniques
have typically been used to address such convex optimization
problems. The first approach (called the batch approach) uses the
ability to evaluate the function $f$ along with its gradients, Hessian
etc. and applies first- and second-order methods to minimize the
objective. The second approach (called the stochastic approach)
interprets the average in Equation~\eqref{eqn:f-defn} as an
expectation and uses stochastic gradient methods, randomly sampling a
$g_i$ and using its gradient and Hessian information as unbiased
estimates for those of the function $f$.\footnote{There is a body of
  literature that recognizes the ability of stochastic optimization to
  minimize testing error rather than training error in machine
  learning contexts~\citep[see e.g.][]{bottou-bousquet-2008}, but we
  will focus on training error for this paper.} Both these classes of
algorithms have extensive literature on upper bounds for the
complexities of specific methods. More fundamentally, there are also
lower bound results on the minimum black-box complexity of the
\emph{best-possible algorithm} to solve convex minimization
problems. In several broad problem classes, these lower bounds further
coincide with the known upper bounds for specific methods, yielding a
rather comprehensive general theory.

However, a recent line of work in the machine learning literature,
recognizes that the specific problem~\eqref{eqn:f-defn} of interest
has additional structure beyond a general convex minimization
problem. For instance, the average in defining the function $f$ is
over a fixed number $n$ of functions, whereas typical complexity
results on stochastic optimization allow for the expectation to be
with respect to a continuous random variable. Recent
works~\citep{RouxScBa2012,Shalev-ShwartzZh2013,JohnsonZh2013} make
further assumptions that the functions $g_i$ involved in this sum are
smooth, and the function $f$ is of course strongly convex by
construction. Under these conditions, the algorithms studied in these
works have the following properties: (i) the cost of each iteration is
identical to stochastic optimization methods, and (ii) the convergence
rate of the method is linear.\footnote{An optimization algorithm is
  linearly convergent if it reduces the sub-optimality by a constant
  factor at each iteration.} The results are surprising since the
existing lower bounds on stochastic optimization dictate that the
error can decrease no faster than $\Omega(1/k)$ after $k$ iterations
under such assumptions~\citep{nemirovsky-yudin-1983}, leaving an
exponential gap compared to these new results. It is of course not a
contradiction due to the finite sum structure of the
problem~\eqref{eqn:f-defn} (following the terminology of
\citet{bertsekas-2010}, we will call the setup of optimizing a finite
sum \emph{incremental optimization} hereafter).

Given this recent and highly interesting line of work, it is natural
to ask just how much better can one do in this model of minimizing
finite sums. Put another way, can we specialize the existing lower
bounds for stochastic or batch optimization, to yield results for this
new family of functions. The aim of such a result would be to
understand the fundamental limits on any possible algorithm for this
family of problems, and whether better algorithms are possible at all
than the existing ones. Answering such questions is the goal of this
work. To this end, we define the Incremental First-order Oracle (IFO)
complexity model, where an algorithm picks an index $i \in
\{1,2,\ldots, n\}$ and a point $x \in \domain$ and the oracle returns
$g_i'(x)$. We consider the setting where each function $g_i$ is
L-smooth (that is, it has L-Lipschitz continuous gradients). In this
setting, we demonstrate that no method can achieve
\mbox{$\|x_K-x^*_f\|\leq\epsilon\|x^*_f\|$} for all functions $f$ of
the form~\eqref{eqn:f-defn}, without performing
\mbox{$K=\Omega\big(n+\sqrt{n\inPar{{L}/{\mu}-1}}\,\log(1/\epsilon)\big)$}
calls to the IFO. As we will discuss following this main result, this
lower bound is not too far from upper bounds for IFO methods such as
SAG, SVRG and SAGA~\citep{SchmidtRoBa2013,JohnsonZh2013,Defazio2014}
whose iteration complexity is $\order((n +
L/\mu)\log(1/\epsilon))$. Some dual coordinate methods such as ASDCA
and SPDC~\citep{Shalev-ShwartzZh2014,ZhangXi2014} get even closer to
the lower bound, but are not IFO algorithms. Overall, there is no
method with a precisely matching upper bound on its complexity,
meaning that there is further room for improving either the upper or
the lower bounds for this class of problems.

Following the statement of our main result, we will also discuss the
implications of these lower bounds for the typical machine learning
problems that have inspired this line of work. In particular, we will
demonstrate that caution is needed in comparing the results between
the standard first-order and IFO complexity models, and worst-case
guarantees in the IFO model might not adequately capture the
performance of the resulting methods in typical machine learning
settings. We will also demonstrate regimes in which different IFO
methods as well as standard first-order methods have their strengths
and weaknesses.

Recent work of~\citet{Arjevani2014} also studies the problem of lower
bounds on smooth and strongly convex optimization methods, although
their development focuses on certain restricted subclasses of
first-order methods (which includes SDCA but not the accelerated
variants, for instance). Discussion on the technical distinctions in
the two works is presented following our main result.

As a prerequisite for our result, we need the result on black-box
first-order complexity of minimizing smooth and strongly convex
functions. We provide a self-contained proof of this result in our
paper in Appendix~\ref{sec:sproblem} which might be of independent
interest. In fact, we establish a slight variation on the original
result, in order to help prove our main result. Our main result will
invoke this construction multiple times to design each of the
components $g_i$ in the optimization problem~\eqref{eqn:f-defn}.

The remainder of this paper is organized as follows. The next section
formally describes the complexity model and the structural
assumptions. We then state the main result, followed by a discussion
of consequences for typical machine learning problems. The proofs are
deferred to the subsequent section, with the more technical details in
the appendix.

\section{Setup and main result}

Let us begin by formally describing the class of functions we will
study in this paper. Recall that a function $g$ is called $L$-smooth,
if it has $L$-Lipschitz continuous gradients, that is
\[
    \forall\: x,y \in \domain \quad \|g'(x)-g'(y)\|_*\leq L \|x-y\|~,
\]
where $\|\cdot\|_*$ is the norm dual to $\|\cdot\|$. In this paper, we
will only concern ourselves with scenarios where $\domain$ is a convex
subset of a separable Hilbert space, with $\|\cdot\|$ being the
(self-dual) norm associated with the inner product. A function $g$ is
called $\mu$-strongly convex if
\[
\forall\: x,y \in \domain \quad g(y) \geq g(x) + \inprod{g'(x)}{y-x} +
\frac{\mu}{2} \|x-y\|^2.
\]
Given these definitions, we now define the family of functions being
studied in this paper.

\begin{definition}
  Let $\calFo$ denote the class of all convex functions $f$ with the
  form~\eqref{eqn:f-defn}, where each $g_i$ is $(L-\mu)$-smooth and
  convex.
  \label{defn-calfo}
\end{definition}
Note that $f$ is $\mu$-strongly convex and $L$-smooth by construction,
and hence $\calFo \subseteq \calSo$ where $\calSo$ is the set of all
$\mu$-strongly convex and $L$-smooth functions. However, as we will
see in the sequel, it can often be a much smaller subset, particularly
when the smoothness of the global function is much better than that of
the local functions. We now define a natural oracle for optimization
of functions with this structure, along with admissible algorithms.

\begin{definition}[Incremental First-order Oracle (IFO)]
  For a function $f \in \calFo$, the Incremental First-order Oracle
  (IFO) takes as input a point $x \in \domain$ and index $i \in
  \{1,2,\ldots, n\}$ and returns the pair $(g_i(x),\,g'_i(x))$.
\end{definition}

\begin{definition}[IFO Algorithm]
An optimization algorithm is an IFO algorithm if its specification does 
not depend on the cost function $f$ other than through calls to an IFO.
\end{definition}
For instance, a standard gradient algorithm
would take the current iterate $x_k$, and invoke the IFO with
$(x_k,i)$ in turn with $i=\{1,2,\ldots, n\}$, in order to assemble the
gradient of $f$. A stochastic gradient algorithm would take the
current iterate $x_k$ along with a randomly chosen index $i$ as inputs
to IFO. Most interesting to our work, the recent SAG, SVRG and SAGA 
algorithms~\citep{RouxScBa2012,JohnsonZh2013,Defazio2014}
are IFO algorithms. On the other hand, dual coordinate ascent algorithms 
require access to the gradients of the conjugate of $f_i$, and therefore are
not IFO algorithms.

We now consider IFO algorithms that invoke the oracle $K$ times (at
$x_0, \ldots, x_{K-1}$) and output an estimate $x_K$ of the minimizer
$x^*_f$.  Our goal is to bound the smallest number of queries $K$
needed for any method to ensure an error $\|x_K -
x^*_f\|\leq\epsilon\|x^*_f\|$, uniformly for all $f \in \calFo$. This
complexity result will depend on the ratio $\kappa=L/\mu$ which is
analogous to the condition number that usually appears in complexity
bounds for the optimization of smooth and strongly convex functions.
Note that $\kappa$ is strictly an upper bound on the condition number
of $f$, but also the best one in general given the structural
information about $f \in \calFo$.

In order to demonstrate our lower bound, we will make a specific
choice of the problem domain~$\domain$. Let $\ell_2$ be the Hilbert
space of real sequences \mbox{$x=(x\of{i})_{i=1}^{\infty}$} with
finite norm \mbox{$\|x\|^2=\sum_{i=1}^{\infty}x\of{i}^2$}, and
equipped with the standard inner product $\inprod{x}{y} =
\sum_{i=1}^\infty x\of{i}y\of{i}$. We are now in a position to state
our main result over the complexity of optimization for the function
class $\calFz$.

\begin{theorem}
\label{thm:fbound}
Consider an IFO algorithm for 
problem~(\ref{eqn:f-defn}) that performs $K\geq0$ calls to the 
oracle and output a solution $x_K$. Then, for any $\gamma>0$, 
there exists a function $f\in\calFz$ such that $\|x^*_f\|=\gamma$ and 
\begin{align*}
\|x^*_f-x_K\| ~\geq~ \gamma q^{2t} \quad &\text{\rm with}~~
q=\frac{\sqrt{1+\frac{\kappa-1}{n}}-1}{\sqrt{1+\frac{\kappa-1}{n}}+1}
~,~ \kappa=\frac{L}{\mu}~~\mbox{and}~~ t = \left\{\begin{array}{ll}0 & \mbox{if
  $K<n$.} \\ K/n & \mbox{otherwise.} \end{array}\right.
\end{align*}
\end{theorem}

In order to better interpret the result of the theorem, we state the
following direct corollary which lower bounds the number of steps need
to attain an accuracy of $\epsilon\|x^*_f\|$.

\begin{corollary}
\label{corr:fbound}
Consider an IFO algorithm for problem~(\ref{eqn:f-defn})
that guarantees  $\|x^*_f -x_K\| \leq \epsilon \|x^*_f\|$ 
for any $\epsilon < 1$. 
Then there is a function \smash{$f\in\calFz$}\rule{0pt}{1.9ex} 
on which the algorithm must perform at least
$K=\Omega(n+\sqrt{n(\kappa-1)}\:\log(1/\epsilon))$ IFO calls.
\end{corollary}

The first term in the lower bound simply asserts that any optimization
method needs to make at least one query per $g_i$, in order to even see
each component of $f$ which is clearly necessary. The second term,
which is more important since it depends on the desired accuracy
$\epsilon$, asserts that the problem becomes harder as the number of
elements $n$ in the sum increases or as the problem conditioning
worsens. Again, both these behaviors are qualitatively
expected. Indeed as $n \rightarrow \infty$, the finite sum approaches
an integral, and the IFO becomes equivalent to a generic
stochastic-first order oracle for $f$, under the constraint that the
stochastic gradients are also Lipschitz continuous. Due to
$\Omega(1/\epsilon)$ complexity of stochastic strongly-convex
optimization (with no dependence on $n$), we do not expect the linear
convergence of Corollary~\ref{corr:fbound} to be valid as $n
\rightarrow \infty$. Also, we certainly expect the problem to get
harder as the ratio $L/\mu$ degrades. Indeed if all the functions
$g_i$ were identical, whence the IFO becomes equivalent to a standard
first-order oracle, the optimization complexity similarly depends on
$\Omega(\sqrt{\kappa - 1}\log(1/\epsilon))$.

Whenever presented with a lower bound, it is natural to ask how it
compares with the upper bounds for existing methods. We now compare
our lower bound to upper bounds for standard optimization schemes for
$\calSz$ as well as specialized ones for $\calFz$. We specialize to
$\|x^*_f\| = 1$ for this discussion.

\paragraph{Comparison with optimal gradient methods:} 
As mentioned before, $\calFz \subseteq \calSz$, and hence standard
methods for optimization of smooth and strongly convex objectives
apply. These methods need $n$ calls to the IFO for getting the
gradient of $f$, followed by an update. Using Nesterov's optimal
gradient method~\citep{nesterov-2004}, one needs at most
$\order(\sqrt{\kappa}\log(1/\epsilon))$ gradient evaluations to reach
$\epsilon$-optimal solution for $f \in \calSz$, resulting in at most
$\order(n\sqrt{\kappa}\log(1/\epsilon))$ calls to the IFO. Comparing
with our lower bound, there is a suboptimality of at most
$\order(\sqrt{n})$ in this result. Since this is also the best
possible complexity for minimizing a general $f \in \calSz$, we
conclude that there might indeed be room for improvement by exploiting
the special structure here. Note that there is an important caveat in
this comparison. For $f$ of the form~\eqref{eqn:f-defn}, the
smoothness constant for the overall function $f$ might be much smaller
than $L$, and the strong convexity term might be much higher than
$\mu$ due to further contribution from the $g_i$. In such
scenarios. the optimal gradient methods will face a much smaller
condition number $\kappa$ in their complexity. This issue will be
discussed in more detail in Section~\ref{sec:ml-opt}.

\paragraph{Comparison with the best known algorithms:}
At least three algorithms recently developed for problem
setting~\eqref{eqn:f-defn} offer complexity guarantees that are close
to our lower bound. SAG, SVRG and
SAGA~\citep{RouxScBa2012,JohnsonZh2013,Defazio2014} all reach an
optimization error~$\epsilon$ after less than
\mbox{$\order((n+\kappa)\log(1/\epsilon))$} calls to the oracle. There
are two differences from our lower bound. The first term of $n$
multiplies the $\log(1/\epsilon)$ term in the upper bounds, and the
condition number dependence is $O(\kappa)$ as opposed to
$O(\sqrt{n\kappa})$. This suggests that there is room to either
improve the lower bound, or for algorithms with a better
complexity. As observed earlier, the ASDCA and SPDC
methods~\citep{Shalev-ShwartzZh2014,ZhangXi2014} reach a closer upper
bound of \mbox{$\order((n+\sqrt{n(\kappa-1)})\log(1/\epsilon))$}, but
these methods are not IFO algorithms. 

\paragraph{Room for better lower bounds?} One natural question to ask is
whether there is a natural way to improve the lower bound. As will
become clear from the proof, a better lower bound is not possible for
the \emph{hard problem instance} which we construct. Indeed for the
quadratic problem we construct, conjugate gradient descent can be used
to solve the problem with a nearly matching upper bound. Hence there
is no hope to improve the lower bounds without modifying the
construction.

It might appear odd that the lower bound is stated in the infinite
dimensional space $\ell_2$. Indeed this is essential to rule out
methods such as conjugate gradient descent solving the problem exactly
in a finite number of iterations depending on the dimension only
(without scaling with $\epsilon$). An alternative is to rule out such
methods, which is precisely the approach~\citet{Arjevani2014}
takes. On the other hand, the resulting lower bounds here are
substantially stronger, since they apply to a broader class of
methods. For instance, the restriction to stationary methods
in~\citet{Arjevani2014} makes it difficult to allow any kind of
adaptive sampling of the component functions $f_i$ as the optimization
progresses, in addition to ruling out methods such as conjugate
gradient. 


\section{Consequences for optimization in machine learning}
\label{sec:ml-opt}

With all the relevant results in place now, we will compare the
efficiency of the different available methods in the context of
solving typical machine learning problems. Recall the definitions of
the constants $L$ and $\mu$ from before. In general, the full
objective $f$~\eqref{eqn:f-defn} has its own smoothness and strong
convexity constants, which need not be the same as $L$ and $\mu$. To
that end, we define $L_f$ to be the smoothness constant of $f$, and
$\mu_f$ to the strong convexity of $f$. It is immediately seen that $L$
provides an upper bound on $L_f$, while $\mu$ provides a lower
bound on $\mu_f$.

In order to provide a meaningful comparison for incremental as well as
batch methods, we follow~\citet{ZhangXi2014} and compare the methods
in terms of their \emph{batch complexity}, that is, how many times one
needs to perform $n$ calls to the IFO in order to ensure that the
optimization error for the function $f$ is smaller than
$\epsilon$. When defining batch complexity, \citet{ZhangXi2014}
observed that the incremental and batch methods have dependence on $L$
versus $L_f$, but did not consider the different strong convexities
that play a part for different algorithms. In this section, we also
include the dual coordinate methods in our comparison since they are
computationally interesting for the problem~\eqref{eqn:f-defn} even
though they are not admissible in the IFO model. Doing so, the batch
complexities can be summarized as in Table~\ref{tbl:compare-general}.

\begin{table}
  \center
  \def\mycite#1{{\goodbreak\raggedright{}\small\citep{#1}}}
  \begin{tabular}{@{~}c@{~~~}c@{~~~}c@{~}} \hline \rule{0pt}{2.2ex}\relax
    Algorithm & Batch complexity & Adaptive? 
    \\ \hline
    \parbox[c]{.35\linewidth}{\flushleft
      ASDCA, SDPC \mycite{Shalev-ShwartzZh2014}\mycite{ZhangXi2014}} 
    & \small{$\otil\inPar{\inPar{1+\sqrt{\frac{L-\mu}{\mu
              n}}}\log\frac{1}{\epsilon}}$}  
    & no 
    \\ 
    \parbox[c]{.35\linewidth}{\flushleft
      SAG \mycite{SchmidtRoBa2013}} 
    & \small{$\otil\inPar{\inPar{1+\frac{L}{\mu_f\,n}}\log\frac{1}{\epsilon}}$ }
    & to $\mu_f$ 
    \\ 
    \parbox[c]{.35\linewidth}{\flushleft
      AGM$^\dagger$ \mycite{Nesterov2007}}
    & \small{$\otil\left(\sqrt{\frac{L_f}{\mu_f}}\log\frac{1}{\epsilon}\right)$}
    & to $\mu_f$, $L_f$ \\[3.5ex] \hline
  \end{tabular}
  \caption{A comparison of the batch complexities of different
    methods. A method is adaptive to $\mu_f$ or $L_f$, if it does not
    need the knowledge of these parameters to run the algorithm and
    obtain the stated complexity upper bound. \quad $^\dagger$Although
    the simplest version of AGM does require the specification of
    $\mu_f$ and $L_f$, Nesterov also discusses an adaptive variant
    with the same bound up to additional logarithmic factors.}
  \label{tbl:compare-general}
\end{table}  

Based on the table, we see two main points of difference. First, the
incremental methods rely on the smoothness of the individual
components. That this is unavoidable is clear, since even the worst
case lower bound of Theorem~\ref{thm:fbound} depends on $L$ and not
$L_f$. As~\citet{ZhangXi2014} observe, $L_f$ can in general be much
smaller than $L$. They attempt to address the problem to some extent
by using non-uniform sampling, thereby making sure that the best of
the $g_i$ and the worst of the $g_i$ have a similar smoothness
constant under the reweighing. This does not fully bridge the gap
between $L$ and $L_f$ as we will show next. However, more striking is
the difference in the lower curvature across methods. To the best of
our knowledge, all the existing analyses of coordinate ascent require
a clear isolation of strong convexity, as in the function
definition~\eqref{eqn:f-defn}. These methods then rely on using $\mu$
as an estimate of the curvature of $f$, and cannot adapt to any
additional curvature when $\mu_f$ is much larger than $\mu$. Our next
example shows this can be a serious concern for many machine learning
problems.

In order to simplify the following discussion we restrict ourselves to
perhaps the most basic machine learning optimization problem, the
regularized least-squares regression:
\begin{equation}
  f(x) = \frac{\mu}{2} \|x\|^2 + \frac{1}{n} \sum_{i=1}^n g_i(x) 
  ~\text{\rm with}~ g_i(x)=\inPar{\inprod{a_i}{x}-\,b_i}^2 ,
  \label{eqn:ml-fdefn}
\end{equation}
where $a_i$ is a data point and $b_i$ is a scalar target for
prediction. It is then easy to see that
$g_i^{\prime\prime}(x)=a_i\,a_i\t$ so that $f\in\calFo$ with
$L=\max_i(\mu+\|a_i\|^2)$. To simplify the comparisons, assume that
$a_i\in\bbbR^d$ are drawn independently from a distribution defined on
the sphere $\|a_i\|=R$. This ensures that $L=\mu+R^2$. Since each
function $g_i$ has the same smoothness constant, the
importance sampling techniques of~\citet{ZhangXi2014} cannot help. 

In order to succinctly compare algorithms, we use the notation
$\Gamma_{\mathrm{ALG}}$ to represent the batch complexity of
$\mathrm{ALG}$ without the $\log(1/\epsilon)$ term, which is common
across all methods. Then we see that the upper bound for
$\Gamma_{\mathrm{ASDCA}}$ is
\begin{equation}
\label{eq:k-asdca}
    \Gamma_{\rm ASDCA} ~=~ 1+\sqrt{\frac{\kappa-1}{n}}~=~ 1 +
    \sqrt{\frac{R^2}{\mu n}}.
\end{equation}

In order to follow the development of Table~\ref{tbl:compare-general}
for SAG and AGM, we need to evaluate the constants $\mu_f$ and $L_f$.
Note that in this special case, the constants $L_f$ and $\mu_f$ are
given by the upper and lower eigenvalues respectively of the matrix
$\mu I+\Sigmahat$, where \mbox{$\Sigmahat=\sum_{i=1}^na_ia_i\t/n$}
represents the empirical covariance matrix.  In order to understand
the scaling of this empirical covariance matrix, we shall invoke
standard results on matrix concentration.

Let \mbox{$\Sigma=\E[a_ia_i\t]$} be the second moment matrix of the
$a_i$ distribution. Let $\lambdamin$ and $\lambdamax$ be its lowest
and highest eigenvalues. Let us define the condition number of the
penalized population objective
\[
   \kappaf ~\stackrel{\Delta}{=}~ \frac{\mu + \lambdamax}{\mu+\lambdamin} ~.
\]


Equation (5.26) in \citep{Vershynin2012} then implies that there
are universal constants $c$ and $C$ such that the following inequality 
holds with probability $1-\delta$\::
\begin{equation*}
    \|\Sigma-\Sigmahat\| ~\leq~ \|\Sigma\|\,\max\inPar{z,z^2}
    ~\mbox{\rm with}~
    z = c\sqrt{\frac{d}{n}}+C\sqrt{\frac{\log(2/\delta)}{n}} ~.
\end{equation*}
Let us weaken the above inequality slightly to use $\mu + \|\Sigma\|$
instead of $\|\Sigma\|$ in the bound, which is minor since we
typically expect $\mu \ll \lambdamax$ for statistical
consistency. Then assuming we have enough samples to ensure that 
\begin{equation}
  c^2\frac{d}{n} + C^2 \frac{\log(d/\delta)}{n} \leq
  \frac{1}{8\kappa_f^2},
\label{eqn:enoughexamples}
\end{equation}
we obtain the following bounds on the eigenvalues of the sample
covariance matrix\::
\begin{eqnarray*}
  &\mu_f \geq
  \max\inBrace{\mu,\:\mu+\lambdamin-\lambdamax\max\inPar{z,z^2}} \geq
  \frac{\mu + \lambdamin}{2}, \\
   &L_f  \leq
   \min\inBrace{L,\:\mu+\lambdamax+\lambdamax\max\inPar{z,z^2}} \leq
   \frac{3(\mu + \lambdamax)}{2}~,
\end{eqnarray*}


Using these estimates in the bounds of Table~\ref{tbl:compare-general} gives
\begin{eqnarray}
  \label{eq:k-sag}
  \Gamma_{\rm SAG} &=& 1 + \frac{L}{\mu_f\,n} 
         ~\leq~ 1 + \frac{2(\mu + R^2)}{n(\mu + \lambdamin)} ~=~ \order(1) \,,~~\quad \\
  \label{eq:k-agm}
  \Gamma_{\rm AGM} &=& \sqrt{\frac{L_f}{\mu_f}}
         ~\leq~ \sqrt{3\kappaf}  ~=~ \order(\sqrt{\kappaf}) ~.
\end{eqnarray}
Table~\ref{tbl:compare-ml} compares the three
methods under assumption~\eqref{eqn:enoughexamples}
depending on the growth of $\kappa$.

\begin{small}
\begin{table}[h]
  \bigskip
  \center
  \begin{tabular}{@{~}c@{~~}cc@{~}} \hline \rule{0pt}{2.2ex}\relax
    Algorithm 
    & $\kappa = \order(n)$ 
    & $\kappa \gg n$ 
    \\\hline \rule{0pt}{3.8ex}\relax 
    ASDCA, SPDC (Eq.~\eqref{eq:k-asdca}) 
    & \small{$\otil\left(\log\frac{1}{\epsilon}\right)$ }
    & \small{$\otil\left(\sqrt{\frac{\kappa}{\rule{0pt}{1.4ex}n}}\log\frac{1}{\epsilon}\right)$}
    \\ \rule{0pt}{3.6ex}\relax
    SAG (Eq.~\eqref{eq:k-sag}) 
    & \small{$\otil\left(  \log\frac{1}{\epsilon} \right)$ }
    & \small{$\otil\left(  \log\frac{1}{\epsilon} \right)$ }
    \\ \rule{0pt}{3.6ex}\relax
    AGM (Eq.~\eqref{eq:k-agm}) 
    & \small{$\otil\left( \sqrt{\kappaf} \log\frac{1}{\epsilon}\right)$ }
    & \small{$\otil\left(  \sqrt{\kappaf} \log\frac{1}{\epsilon}\right)$ }
    \\[1ex] \hline
  \end{tabular}
  \caption{A comparison of the batch complexities of different methods
    for the regularized least squares objective~\eqref{eqn:ml-fdefn}
    when the number of examples is sufficiently
    large~\eqref{eqn:enoughexamples}. Observe how the ASDCA complexity
    bound can be significantly worse than the SAG complexity bound, 
    despite its better worst case guarantee.}
  \label{tbl:compare-ml}
\end{table}
\end{small}
\goodbreak

\paragraph{Problems with $\kappa = \order(n)$:} 
This setting is quite interesting for machine learning, since it
corresponds roughly to using $\mu = \order(n)$ when $R^2$ is a
constant. In this regime, all the incremental methods seem to enjoy
the best possible convergence rate of $\otil(\log(1/\epsilon))$. When
the population problem is relatively well conditioned, AGM obtains a
similar complexity since $\kappa_f = \order(1)$. However, for poorly
conditioned problems, the population condition number might scale with
the dimension $d$. We conclude that there is indeed a benefit from
using the incremental methods over the batch methods in these
settings, but it seems hard to distinguish between the complexities of
accelerated methods like ASDCA and SPDC compared with SAG or SVRG.

\paragraph{Problems with large $\kappa$:} In this setting, the
coordinate ascent methods seem to be at a disadvantage, because the
average loss term provides additional strong convexity, which is
exploited by both SAG and AGM, but not by ASDCA or SPDC
methods. Indeed, we find that the complexity term $\Gamma_{\rm ASDCA}$
can be made arbitrarily large as $\kappai$ grows large. However, the
contraction factors for both SAG and AGM do not grow with $n$ in this
setting, leading to a large gap between the complexities. Between SAG
and AGM, we conclude that SAG has a better bound when the population
problem is poorly conditioned. 

\paragraph{High-dimensional settings (\boldmath $n/d\ll1$)\::} In this
setting, the global strong convexity can not really be larger than
$\mu$ for the function~\eqref{eqn:ml-fdefn}, since the Hessian of the
averaged loss has a non-trivial null space. It would appear then, that
SAG is forced to use the same problem dependent constants as
ASDCA/SPDC, while AGM gets no added benefit in strong convexity
either. However, in such high-dimensional problems, one is often
enforcing a low-dimensional structure in machine learning settings for
generalization. In such structures, the global Hessian matrix can
still satisfy restricted versions of strong convexity and smoothness
conditions, which are often sufficient for batch optimization methods
to succeed~\citep{AgarwalNeWa2012}. In such situations, the comparison
might once again resemble that of Table~\ref{tbl:compare-ml}, and we
leave such development to the reader.

In a nutshell, the superiority of incremental algorithms for the
optimization of training error in machine learning is far more subtle
than suggested by their worst case bounds.  Among the incremental
algorithms, SAG has favorable complexity results in all regimes
despite the fact that both ASDCA and SPDC offer better worst case
bounds. This is largely due to the adaptivity of SAG to the curvature
of the problem. This might also explain in some part the empirical
observation of~\citet{SchmidtRoBa2013}, who find that on some datasets
SDCA (without acceleration) performed significantly poorly compared
with other methods (see Figure 2 in their paper for details). Finally,
we also observe that SAG does indeed improve upon the complexity of
AGM after taking the different problem dependent constants into
account, when the population problem is ill-conditioned and the data
are appropriately bounded.

It is worth observing that all our comparisons are ignoring constants,
and in some cases logarithmic factors, which of course play a role in
the running time of the algorithms in practice. Note also that the
worst case bounds for the incremental methods account for the worst
possible choice of the $n$ functions in the sum. Better results might
be possible when they are based on i.i.d. random data. Such results
would be of great interest for machine learning.



\section{Proof of main result}

In this section, we  provide the proof of
Theorem~\ref{thm:fbound}. Our high-level strategy is the following. We
will first construct the function $f~:~\ell_2\mapsto \bbbR$ such that
each $g_i$ acts on only the projection of a point $x$ onto a smaller
basis, with the bases being disjoint across the $g_i$. Since the $g_i$
are separable, we then demonstrate that optimization of $f$ under an
IFO is equivalent to the optimization of each $g_i$ under a standard
first-order oracle. The functions $g_i$ will be constructed so that
they in turn are smooth and strongly convex with appropriate
constants. Hence, we can invoke the known result for the optimization
of smooth and strongly convex objectives under a first-order oracle,
obtaining a lower bound on the complexity of optimizing $f$. We will
now formalize this intuitive sketch.

\subsection{Construction of a separable objective}

We start with a simple definition.

\begin{definition}
Let $e_1,e_2,\dots$ denote the canonical basis vectors of $\ell_2$,
and let $Q_i$, $i=1\dots n$, denote the orthonormal families
$Q_i\:=\:\inBrack{\:e_i,\,e_{n+i},\,e_{2n+i},\dots,\,e_{kn+i},\dots }$~.
\end{definition}

For ease of presentation, we also extend the transpose notation for
matrices over operators in $\ell_2$ in the natural manner (to avoid
stating adjoint operators each time).

\begin{definition}
Given a finite or countable orthonormal 
family $S=[s_1,s_2,\dots]\subset\ell_2$ and $x\in\ell_2$, let
\[
     S\,x~{=}~\sum_{i=1}^\infty x\of{i}\,s_i  \quad\text{\rm and}\quad
     S\t x~{=}~( \inprod{s_i}{x} )_{i=1}^{\infty}~,
\]
where $s_i$ is assumed to be zero when $i$ is greater than the size of
the family.
\label{defn:orthonormal}
\end{definition}

\begin{remark}
Both $S\,x$ and $S\t x$ are square integrable and therefore belong to
$\ell_2$. 
\end{remark}

Using the above notation, we first establish some simple identities
for the operators $Q_i$ defined above.

\begin{lemma}
Simple calculus yields the following identities:
\begin{align*}
  Q_i\t\,Q_i = \sum_{i=1}^{n} Q_i\,Q_i\t = I,~\textrm{and}~ 
   \|Q_i\,x\|^2 = \sum_{i=1}^{n} \|Q_i\t\,x\|^2 = \|x\|^2~.
\end{align*}
\label{lemma:basis-ident}
\end{lemma}

\begin{proof*}
  We start with the first claim. For any basis vector $e_j$, it is
  easily checked that $Q_ie_j = e_{(j-1)n + i}$. By definition of
  $Q_i\t$, it further follows that $Q_i\t e_{(j-1)n + i} =
  e_j$. Linearity now yields $Q_i\t\,Q_ix = x$ for any $x \in
  \ell_2$, giving the first claim. For the second claim, we observe
  that $Q_iQ_i\t\, e_j = 0$ unless $\textrm{mod}(j,n) = i$, in which case
  $Q_iQ_i\t\, e_j = e_j$. This implies the second claim. The third
  claim now follows from the first one, since $\inprod{Q_i\,x}{Q_i\,x}
  = \inprod{x}{Q_i\t\,Q_i\,x} = \|x\|^2$. Similarly the final claim
  follows from the second claim. 
\end{proof*}

We now define the family of separable functions 
that will be used to establish our lower bound. 
\begin{equation}
\label{eq:separated}
    f(x) = \frac{\mu}{2} \|x\|^2 + \frac{1}{n}\sum_{i=1}^{n} h_i(Q_i\t  x) 
           ~,~~~  h_i(x)\in\calS^{0,L-\mu}(\ell_2) ~~
\end{equation}

\begin{proposition}
All functions (\ref{eq:separated}) belong to $\calFz$.
\end{proposition}
\begin{proof*}
We simply need to prove that the functions $g_i(x)=h_i(Q_i\t x)$
belong to $\calS^{0,L-\mu}(\ell_2)$.  Using
$g^\prime_i(x)=Q_i\,h^\prime_i(Q_i\t x)$ and
Lemma~\ref{lemma:basis-ident}, we can write $\displaystyle \|
g^\prime_i(x)-g^\prime_i(y) \|^2 = \big\| Q_i\,\big(\,h^\prime_i(Q_i\t
x)-h^\prime_i(Q_i\t y)\,\big)\,\big\|^2 = $ $\displaystyle \|
h^\prime_i(Q_i\t x)-h^\prime_i(Q_i\t y) \|^2 \leq (L-\mu)^2
\|Q_i\t(x-y)\|^2 \leq (L-\mu)^2 \|x-y\|^2 ~.$
\end{proof*}

\subsection{Decoupling the optimization across components}

We would like to assert that the separable structure of $f$ allows
us to reason about optimizing its components separately. 
Since the $h_i$ are not strongly convex by themselves, we first
rewrite $f$ as a sum of separated strongly convex functions.
Using Lemma~\ref{lemma:basis-ident},
\begin{align*}
 f(x) &=  \frac{\mu}{2} \|x\|^2 + \frac{1}{n}\sum_{i=1}^{n}
   h_i(Q_i\t x) = \frac{\mu}{2} \sum_{i=1}^{n}\|Q_i\t
 x\|^2+\frac{1}{n}\sum_{i=1}^{n}h_i(Q_i\t x)\\ 
 &= \frac{1}{n}\sum_{i=1}^{n}\inBrack{\frac{n\mu}{2}\|Q_i\t x\|^2+h_i(Q_i\t x)}
 ~\stackrel{\Delta}{=}~ \frac{1}{n}\sum_{i=1}^{n} f_i(Q_i\t x) ~,
\end{align*}
By construction,
the functions $f_i$ belong to $\calS^{n\mu,L-\mu+n\mu}$ and are
applied to disjoint subsets of the $x$ coordinates.  Therefore, when
the function is known to have form~(\ref{eq:separated}),
problem~(\ref{eqn:f-defn}) can be written as
\begin{equation}
\label{eq:fiproblems}
  x^* = \sum_{i=1}^n Q_i x^*_i \qquad x^*_i ~=~ \argmin_{x\in\ell_2} ~
  f_i(x)~.
\end{equation}
Any algorithm that solves optimization problem~(\ref{eqn:f-defn})
therefore implicitly solves all the problems listed
in~(\ref{eq:fiproblems}).

We are almost done, but for one minor detail. Note that we want to
obtain a lower bound where the IFO is invoked for a pair $(i,x)$ and
responds with $h_i(Q_i\t x)$ and $\partial h_i(Q_i\t x)/\partial x$. 
In order to claim that this suffices to optimize each $f_i$ separately,
we need to argue that a first-order oracle for $f_i$ can be obtained
from this information, knowing solely the structure of $f$ and not the
functions $h_i$. Since the strong convexity constant $\mu$ is assumed
to be known to the algorithm, the additional $(n\mu/2)\|x\|^2$ in
defining $f_i$ is also known to the algorithm. As a result, given an IFO 
for $f$, we can construct a first-order oracle for any of the
$f_i$ by simply returning $h_i(Q_i\t x) + (n\mu/2)\|Q_i\t x\|^2$ and $\partial
h_i(Q_i\t x)/\partial x + n\mu Q_iQ_i\t x)$. Furthermore, an IFO
invoked with the index $i$ reveals no information about $f_j$ for
any other $j$ based on the separable nature of our problem. Hence, the
IFO for $f$ offers no additional information beyond having a standard
first-order oracle for each $f_i$. 

\subsection{Proof of Theorem~\ref{thm:fbound}}

Based on the discussion above, we can pick any $i\in\{1\dots n\}$
and view our algorithm as a complicated setup whose sole purpose is to
optimize function $f_i\in\calS^{n\mu,L-\mu+n\mu}$. Indeed, given the
output $x_K$ of an algorithm using an IFO for the function $f$, we can
declare $x^i_K = Q_i\t x_K$ as our estimate for
$x^*_i$. Lemma~\ref{lemma:basis-ident} then yields
\begin{align*}
  \|x_K - x^*_f\|^2 &= \sum_{i=1}^n \|Q_i\t (x_K - x^*_f)\|^2 =
  \sum_{i=1}^n \|x^i_K - x^*_i\|^2. 
\end{align*}
In order to establish the theorem, we now invoke the classical result
on the black-box optimization of functions using a first-order
oracle. The specific form of the result stated here is proved in
Appendix~\ref{sec:sproblem}. 

\begin{theorem}[Nemirovsky-Yudin]
\label{thm:sbound}
Consider a first order black box optimization algorithm for
problem~(\ref{eq:sproblem}) that performs $K\geq0$ calls to 
the oracle and returns an estimate $x_K$ of the minimum. 
For any $\gamma>0$, there exists a function $f\in\calSz$ such that
$\|x^*_f\|=\gamma$ and 
\[
   \|x^*_f - x_K\| ~\geq~ \gamma \:\: q^{2K} \quad
     \text{\rm with}~~ q=\frac{\sqrt{\kappa}-1}{\sqrt{\kappa}+1} ~~
          \text{\rm and~~} \kappa=\frac{L}{\mu}~.
\]
\end{theorem}

At a high-level, our oracle will make an independent choice of one of
the functions that witness the lower bound in Theorem~\ref{thm:sbound}
for each $f_i$. At a high-level, each function $f_i$ will be chosen to
be a quadratic with an appropriate covariance structure such that
$K_i$ queries to the function $f_i$ result in the estimation of at
most $K_i+1$ coordinates of $x_i^*$. By ensuring that the remaining
entries still have a substantial norm, a lower bound for such
functions is immediate. The precise details on the construction of
these functions can be found in
Appendix~\ref{sec:sproblem}.\footnote{The main difference with the
  original result of Nemirovsky and Yudin is the dependence on
  $\gamma$ instead of $\|x_0 - x^*_f\|$. This is quite convenient in
  our setting, since it eliminates any possible interaction amongst
  the starting values of different coordinates for the different
  functions $f_i$.}

Suppose the IFO is invoked $K_i$ times on each index $i$, 
with $K = K_1 + K_2 + \ldots + K_n$. We first establish the theorem 
for the case $K < n$ in which the algorithm cannot query each 
functions $f_i$ at least once. After receiving the response $x_K$,
we are still free to arbitrarily choose $f_i$ for any index $i$
that was never queried.  No non-trivial accuracy is possible
in this case.

\begin{proposition}
  Consider an IFO algorithm that satisfies the conditions 
  of Theorem~\ref{thm:fbound} with $K<n$.
  Then there is a function $f \in \calFz$ such that 
  $\|x^*_f-x_K\| ~\geq~ \gamma$.
  \label{prop:smallK}
\end{proposition}

\begin{proof*}
  Let us execute the algorithm assuming that all the $f_i$
  are equal to the function $f$ of Theorem~\ref{thm:sbound} that 
  attains the lower bound with $\gamma = 0$. Since $K<n$,
  there is at least one function $f_j$ for which $K_j=0$.
  Since the IFO has not revealed anything about this function,
  we can construct 
  function $f$ by redefining function $f_j$ to ensure that
   $\|x^j_K - x^*_j\| \geq \|x^*_j\| = \gamma_j$. 
   Since $x^*_j$ is the only part of $x^*$
  which is non-zero, we also get $\gamma_j = \gamma$.
\end{proof*}

We can now assume without loss of generality that $K_i> 0$ for each $i$. 
Appealing to Theorem~\ref{thm:sbound} for each $f_i$ in turn,
\begin{align*}
  \|x_K - x^*_f\|^2 ~~=~~ \sum_{i=1}^n \|x^i_K - x^*_i\|^2 ~~\geq~~
  \sum_{i=1}^n \gamma_i^2 q^{4K_i}
  ~~=~~ \gamma^2\sum_{i=1}^n \frac{\gamma_i^2}{\gamma^2} q^{4K_i}
  ~~\geq~~ \gamma^2 q^{\sum_{i=1}^n \gamma_i^2 4K_i/\gamma^2},
\end{align*}
where the last inequality results from Jensen's inequality
applied to the convex function $q^{4\alpha}$ for $\alpha \geq 1$. 
Finally, since the oracle has no way to discriminate
amongst the $\gamma_i$ values when $K_i > 0$,  it will end up
setting $\gamma_i = \gamma/\sqrt{n}$. With this setting, we now obtain
the lower bound 
\begin{align*}
  \|x_K - x^*_f\|^2 &\geq \gamma^2 q^{4K/n},
\end{align*}
for $K > n$, along with $\|x_K - x^*_f\|^2 \geq \gamma^2$ for $K <
n$. 

This completes the proof of the Theorem. In order to further establish
Corollary~\ref{corr:fbound}, we need an additional technical lemma.

\begin{lemma}
\label{lemma:magicbound}
$\displaystyle
   \forall x>1 ~,~~
     \log\inPar{\frac{\sqrt{x}-1}{\sqrt{x}+1}} 
       \: > \: \frac{-2}{\sqrt{x-1}} ~.
$
\end{lemma}
\begin{proof*}
  The function
  $\phi(x)=\log\inPar{\frac{\sqrt{x}-1}{\sqrt{x}+1}}+\frac{2}{\sqrt{x-1}}$
  is continuous and decreasing on $(1,+\infty)$ because
  \begin{align*}
   \phi'(x) &= \frac{1}{(\sqrt{x}-1)(\sqrt{x}+1)\sqrt{x}} -
   \frac{1}{(x-1)\sqrt{x-1}}\\  &=~ \frac{1}{(x-1)\sqrt{x}} -
   \frac{1}{(x-1)\sqrt{x-1}} < 0 ~. 
  \end{align*}
  The result follows because $\lim_{x\rightarrow\infty}\phi(x)=0$.
\end{proof*}

Now we observe that we have at least $n$ queries due to the
precondition $\epsilon < 1$ and
Proposition~\ref{prop:smallK}, which yields the first term in the
lower bound. Based on Theorem~\ref{thm:fbound} and this lemma, the
corollary is now immediate.


\section{Discussion} 

The results in this paper were motivated by recent results and
optimism on exploiting the structure of minimizing finite sums, a
problem which routinely arises in machine learning. Our main result
provides a lower bound on the limits of gains that might be possible
in this setting, allowing us to do a more careful comparison of this
setting with regular first-order black box complexity results. As
discussed in Section~\ref{sec:ml-opt}, the results seem mixed when the
sum consists of $n$ functions based on random data drawn i.i.d. from a
distribution. In this statistical setting, we find that the worst-case
near-optimal methods like ASDCA can often be much worse than other
methods like SAG and SVRG. However, IFO methods like SAG certainly
improve upon optimal first-order methods agnostic of the finite sum
structure, in ill-conditioned problems. In general, we observe that
the problem dependent constants that appear in different methods can
be quite different, even though this is not always recognized. We
believe that accounting for these opportunities might open door to
more interesting algorithms and analysis.

Of course, there is another and a possibly more important aspect of
optimization in machine learning which we do not study in this
paper. In typical machine learning problems, the goal of optimization
is not just to minimize the objective $f$---usually called the
training error---to a numerical precision. In most problems, we
eventually want to reason about test error, that is the accuracy of
the predictions we make on unseen data. There are existing
results~\citep{bottou-bousquet-2008} which highlight the optimality of
\emph{single-pass stochastic gradient} optimization methods, when test
error and not training error is taken into consideration. So far, we
do not have any clear results comparing the efficacy of methods
designed for the problem~\eqref{eqn:f-defn} in minimizing test error
directly. We believe this is an important question for future
research, and one that will perhaps be most crucial for the
adoption of these methods in machine learning. 

We believe that there are some important open questions for future
works in this area, which we will conclude with:

\begin{enumerate}
\item Is there a fundamental gap between the best IFO methods and the
  dual coordinate methods in the achievable upper bounds? Or is there
  room to improve the upper bounds on the existing IFO methods. We
  certainly found it tricky to do the latter in our own attempts.
\item Is it possible to obtain better complexity upper bounds when the
  $n$ functions involved in the sum~\eqref{eqn:f-defn} are based on
  random data, rather than being $n$ arbitrary functions? Can the
  incremental methods exploit global rather than local smoothness
  properties in this setting?
\item What are the test error properties of incremental methods for
  machine learning problems? Specifically, can one do better than just
  adding up the optimization and generalization errors, and follow a
  more direct approach as the stochastic optimization literature?
\end{enumerate}


\section*{Acknowledgements}


We would like to thank Lin Xiao, Sham Kakade and Rong Ge for helpful
discussions regarding the complexities of various methods. We also
thank the anonymous reviewer who pointed out that the dual coordinate
are not valid IFO algorithms.



\bibliography{finitesums}

\newpage

\onecolumn
\appendix


\section{Optimization of a strongly convex smooth functions}
\label{sec:sproblem}

The most accessible derivation of this classic lower bound
\cite{nesterov-2004} relies on the simplifying assumption that the
successive points~$x_k$ lie in the span of the gradients previously
returned by the oracle. This section provides a derivation of the
lower bound that does not rely on this assumption and is critical for
Theorem~\ref{thm:fbound} where no such assumptions are made.

This section considers algorithms that produces an
approximate solution of the optimization problem
\begin{equation}
\label{eq:sproblem}
    x^*_f ~=~ \argmin_{x\in\ell_2} ~ f(x) ~ = ~ \frac{\mu}{2} \|x\|^2 + g(x)
    \qquad\text{\rm where}~~ f(x)\in\calSz ~.
\end{equation}
using, as sole knowledge of function $f$, an oracle that returns the
value~$f(x)$ and the gradient $f'(x)$ on points successively determined by the
algorithm. Note that this writing of $f$ is without loss of
generality, since any $\mu$-strongly convex function can be written in
the form~\eqref{eq:sproblem} where $g$ is convex. 

\begin{remark}
\label{remark:foracle-vs-goracle}
We could equivalently consider an oracle that reveals $g(x_k)$ and $g'(x_k)$
instead of $f(x_k)$ and $f'(x_k)$ because these quantities can be computed
from each other (since $\mu$ is known.)
\end{remark}

At a high-level, our proof will have the following structure. We will
first establish that any algorithm for solving the minimization
problem~\eqref{eq:sproblem} for all $f \in \calSz$ will be forced to
play the point $x_K$ in the span of the previous iterates and
gradients. This essentially shows that the restriction made by
Nesterov is not too serious. The second part of the proof constructs a
resisting oracle for such algorithms whose final query point falls
within the span of the previous responses. Combining these
ingredients, we obtain the desired lower bound.

\subsection{Restriction of final solution to span}

Consider an algorithm that calls the oracle on $K>1$ successive points
$x_0,\dots,x_{K-1}$. The first part of the proof describes how to pick
the best possible~$x_K$ on the basis of the oracle answers and the
algorithm's queries.

\begin{definition}
For any $\gamma\geq0$, 
let $\calSzg$ be the set of all functions $f\in\calSz$ that reach
their minimum in a point $x^*_f$ such that $\|x^*_f\|=\gamma$.
\end{definition}

\begin{definition}
Let $\calG_\gamma^f\subset\calSzg$ be the set of the functions of
$\calSzg$ whose values and gradients coincide with those of $f$ on
points $x_0\dots x_{K-1}$.  Let $H_\gamma^f\in\ell_2$ be the set of
their minima.
\end{definition}

\medskip
When the function $f$ is clear from the context, we will drop the
superscript for brevity. Since all functions in $\calG_\gamma$ are
compatible with the values returned by the oracle, running our
algorithm on any of them would perform the same calls to the oracle
and obtain the same answers. Therefore, in order to offer the best
guarantee on $\|x_K-x^*_f\|^2$ without further knowledge of the
function $f(x)$, our algorithm must choose $x_K$ to be the center of
the smallest ball containing $H_\gamma$.


\begin{definition}
Let $P=\Span\inBrace{x_0\dots\,x_{K-1},\,f'(x_0)\dots\,f'(x_{K-1})}$. 
Let $\Pi_P(x)$ be the orthogonal projection of point $x$ on $P$ and
let $M_p(x)=2\Pi_P(x)\!-\!x$ be its mirror image with respect to $P$.
\end{definition}

Stated differently, we know that any point $x$ can be decomposed into
$\Pi_P(x)$ and $\Pi_{P^\perp}(x)$ such that $x = \Pi_P(x) +
\Pi_{P^\perp}(x)$. Then the above definition yields $M_P(x) = \Pi_P(x)
- \Pi_{P^\perp}(x)$, which is the natural reflection of $x$ with
respect to the subspace $P$. 

\begin{proposition}
\label{prop:symmetry}
The set $H_\gamma$ is symmetric with respect to $P$.
\end{proposition}
\begin{proof*}
Consider an arbitrary point $x^*_h\in H_\gamma$ which minimizes a function
$h\in\calG_\gamma$. Since function $x\mapsto h(M_P(x))$ also belongs to
$\calG_\gamma$, its minimum $M_p(x^*_h)$ also belongs to $H_\gamma$.
\end{proof*}
\begin{corollary}
\label{cor:center}
The center of the smallest ball enclosing $H_\gamma$ belongs to $P$.
\end{corollary}

We are now in a position to present the main ingredient of our proof
that allows us to state a more general result than Nesterov. In
particular, we demonstrate that the assumption made by Nesterov about
the iterates lying in the span of previous gradients can be made
almost without loss of generality. The key distinction is that we can
only make it on the step $K$, where the algorithm is constrained to
produce a good answer, while Nesterov assumes it on all iterates,
somewhat restricting the class of admissible algorithms.

\begin{lemma}
\label{lemma:span}
For any $\gamma>0$ and any algorithm $\algA$ that performs $K\geq1$
calls of the oracle and produces an approximate solution
$x^\algA_K(f)$ of problem~(\ref{eq:sproblem}), there is an algorithm
$\algB$ that performs $K$ calls of the oracle and produces an
approximate solution $x^\algB_K(f)
\:\in\:\Span\inBrace{x_0\dots\,x_{K-1},\,f'(x_0)\dots\,f'(x_{K-1})}$
for all $f \in \calSzg$ such that
\begin{equation}
\label{eq:asgoodguarantee}
   \sup_{f\in\calSzg} \|x^\algB_K-x^*_f\|^2 ~\leq~ 
                \sup_{f\in\calSzg} \|x^\algA_K-x^*_f\|^2 ~.
\end{equation}
\end{lemma}
\begin{proof*}
Consider an algorithm $\algB$ that first runs algorithm $\algA$ and then
returns the center of the smallest ball enclosing $H_\gamma^f$ as
$x^\algB_K(f)$. Corollary~\ref{cor:center} ensures that $x^\algB_K(f)$ 
belongs to the posited span. This choice of $x^\algB_K(f)$ also ensures that 
\mbox{$ \sup_{\bar{x}\in H_\gamma^f} \|x^\algB_K(f) - \bar{x}\| \leq
   \sup_{\bar{x}\in H_\gamma^f} \|x^\algA_K(f) - \bar{x}\| $.}
Equivalently,
\mbox{$ \sup_{g\in G_\gamma^f} \|x^\algB_K(g) - x^*_g\| \leq
   \sup_{g\in G_\gamma^f}  \|x^\algA_K(g) - x^*_g\| $,}
where we use the fact that $x^\algB_K(g)=x^\algB_K(f)$ and
$x^\algA_K(g)=x^\algA_K(f)$ because function $g\in G_\gamma^f$ coincides
with $f$ on $x_0\dots x_{K-1}$. Therefore,
\[ \sup_{f\in\calSzg}~\sup_{g\in G_\gamma^f} \|x^\algB_K(g) - x^*_g\| ~\leq~
   \sup_{f\in\calSzg}~\sup_{g\in G_\gamma^f}  \|x^\algA_K(g) - x^*_g\| ~. \]
This inequality implies \eqref{eq:asgoodguarantee}
because $\calSzg=\cup_{f\in\calSzg} G_\gamma^f$.
\end{proof*}

\bigskip
Lemma~\ref{lemma:span} means that we can restrict the analysis to
algorithms that pick their final estimate $x_K$ in the subspace $P$
that results from the execution of the algorithm. In order to
establish a lower bound for such an algorithm, it is sufficient to
construct a function $f_K$ whose minimum is located sufficiently far
away from this subspace.  We construct this function by running the
algorithm against a \emph{resisting oracle}, which is quite standard
in these lower bound proofs. Each call to the resisting oracle picks a
new objective function $f_k$ among all the $\calSz$ functions that
agree with the values and gradients returned by all previous calls to
the oracle. This constraint ensures that the optimization algorithm
would have reached the same state if it had been run against function
$f_k$ instead of the resisting oracle.

\subsection{Construction of a resisting oracle}

We start by defining the basic structure of the function which will be
used by our oracle to construct hard problem instances. This structure
is identical to that used by Nesterov.

\begin{definition}[Nesterov]
Fix $\rho>0$ and let $N_{\mu,L}$ denote the function
\[
   N_{\mu,L}(x) = 
    \frac{L-\mu}{8}\inPar{(x\of1)^2+\sum_{i=1}^{\infty}(x\of{i+1}-x\of{i})^2
                           -2\,\rho\,x\of1} + \frac{\mu}{2} \|x\|^2 ~.
\]
\end{definition}

\begin{proposition}
\label{prop:minimum}
$N_{\mu,L}\in\calSz$ and
reaches its minimum in $x^*_N =  (\rho\,q^i)_{i=1}^{\infty}$
with $q=\frac{\sqrt{\kappa}-1}{\sqrt{\kappa}+1}$.
\end{proposition}
\begin{proof*}
The assertions $\mu I \preceq N^\pprime_{\mu,L} \preceq L I $ and
$N^\prime_{\mu,L}(x^*_N)=0$ follow from direct calculation, as shown
in~\citet[p.\:67]{nesterov-2004}.
\end{proof*}
\begin{remark}
We can arbitrarily choose the value of $\|x^*_N\|$ by appropriately
selecting $\rho$.
\end{remark}

We also need some other properties of the function, which are also
present in Nesterov's analysis. 

\begin{proposition}
\label{prop:increasebyone}
Let $[e_1,e_2,\dots]$ be the canonical basis of $\ell_2$
and let $R_k=\Span(e_1\dots e_{k})$. 
\[
   x\in R_k ~~\Rightarrow~~ N^\prime_{\mu,L}(x)\in R_{k+1} ~.
\]
\end{proposition}
\begin{proof*}
Through a direct calculation, it is easy to verify that  
\begin{align*}
  \frac{\partial}{\partial x\of{i}} N_{\mu,L}(x) &=
  \left\{ \begin{array}{cc} \frac{L -
    \mu}{4} \left(x\of{1} + (x\of{1} - x\of{2} - 2\rho \right) + \mu
  x\of{1} & \mbox{for $i=1$},\\
  \frac{L-\mu}{4}(2x\of{i} - x\of{i+1} - x\of{i-1}) & \mbox{for $i >
    1$}. \end{array}
  \right. 
\end{align*}
The statement directly follows from this. 
\end{proof*}

We now recall our earlier definition of the matrix notation for
orthonormal families in Definition~\ref{defn:orthonormal}. The
resisting oracle we construct will apply the function $N_{\mu,L}$ to
appropriately rotated versions of the point $x$, that is, it constructs
functions of the form $N_{\mu, L}(S\t x)$, where the orthonormal
operators $S$ will be constructed appropriately to ensure that the
optimal solution is sufficiently far away from the span of algorithm's
queries and the oracle's responses. Before we define the oracle, we
need to define the relevant orthogonalization operations.

\begin{definition}[Gram-Schmidt]
Given a finite orthonormal family $S$ and a vector $v$,
the Gram-Schmidt operator $\mathrm{Gram}(S,v)$ augments the
orthonormal family, ensuring that $v$ lies in the new span.
\[
   \mathrm{Gram}(S,v) = \left\{\begin{array}{cl}
      S & \text{\rm if~} v\in\Span(S) \\
      \inBrack{S, \frac{v\,-\,S\,S\t v}{\|v\,-\,S\,S\t v\|}} & \text{\rm otherwise}
   \end{array}\right.
\]
\end{definition}

Our resisting oracle incrementally constructs orthonormal families $S_k$ and
defines the functions $f_k(x)$ as the application of function $N_{\mu,L}$ to
the coordinates of $x$ expressed an orthonormal basis of $\ell_2$ constructed
by completing~$S_k$.
\begin{definition}[Resisting oracle]
Let $S_{-1}$ be an empty family of vectors. Each call $k=0\dots{K\!-\!1}$ 
of the resisting oracle performs the following computations and returns
$y_k=f_k(x_k)$ and $g_k=f^\prime_k(x_k)$.
  \begin{eqnarray}
     \label{eq:ro:span}
       S_k & = & \mathrm{Gram}( \mathrm{Gram}(S_{k-1}, x_k), v_k) 
            \quad \text{\rm for some $v_k\notin\Span(S_{k-1},x_k)$.} \\
     \label{eq:ro:complete}
       \bar{S}_k & = & \inBrack{S_k,\dots} \\
     \label{eq:ro:value}
       y_k & = & f_k(x_k)\, ~=~ N_{\mu,L}(\bar{S}_k\t x_k) \\
     \label{eq:ro:gradient}
       g_k & = & f^\prime_k(x_k) ~=~ \bar{S}_k\:N^\prime_{\mu,L}(\bar{S}_k\t x_k)
  \end{eqnarray}
\end{definition}

\bigskip
Step~(\ref{eq:ro:span}) augments $S_{k-1}$ to ensure that $\Span(S_k)$
contains both $x_k$ and an arbitrary additional vector. This construction
ensures that $\dim(S_k)\leq 2k+2$.
Step~(\ref{eq:ro:complete}) nominally constructs an orthonormal basis
$\bar{S_k}$ of $\ell_2$ by completing $S_k$. This is mostly for notational
convenience because the additional basis vectors have no influence on the
results produced by oracle.
Step~(\ref{eq:ro:value}) computes the value of $y_k=f_k(x_k)$ by applying the
function $N_{\mu,L}$ to the coordinates $\bar{S}_k\t\,x_k$ of
vector~$x_k$ in basis~$\bar{S}_k$. Since $x_k$ belongs to the span of the
first $\dim(S_k)\!-\!1$ basis vectors,
\mbox{$\bar{S}_k\t\,x_k\in{R_{\dim(S_k)-1}}$}.
Finally, step~(\ref{eq:ro:gradient}) computes the gradient
$g_k=f^\prime_k(x_k)$.  Note that $g_k\in S_k$ because
proposition~\ref{prop:increasebyone} ensures that
{$N^\prime_{\mu,L}(\bar{S}_k\t x_k)\in{R_{\dim(S_k)}}$}.

\begin{proposition}
The resisting oracle satisfies the following properties:
\begin{itemize}
\item[(a)~~]   
  $S_k=\Span\inBrace{x_0\dots\,x_{K-1},\,f'(x_0)\dots\,f'(x_{K-1})}
   \quad \dim(S_k)\leq 2k+2$ ~.
\item[(b)~~] 
  $\forall~i<k \quad y_i=f_k(x_i) \quad g_i=f^\prime_k(x_i)$ ~.
\end{itemize}
\end{proposition}
\begin{proof*}
Property (a) holds by construction (see discussion above). Property (b) holds
because both $x_i$ and $g_i$ belong to $\Span(S_i)$. Therefore $y_i=f_k(x_i)$
because $S_i\t x_i = S_k\t x_i$ and $g_i=f^\prime_k(x_i)$ because
\mbox{$N^\prime_{\mu,L}(\bar{S}_k\t x_i)=N^\prime_{\mu,L}(\bar{S}_i\t x_i)
\in{R_{\dim(S_i)}}$}.
\end{proof*}

\subsection{Proof of Theorem~\ref{thm:sbound}}

We now have all the ingredients to establish the main result of this
appendix on the complexity of optimizing smooth and strongly convex
functions. Given our work so far, we know that the solution $x_K$
lives in a $2K+2$ dimensional subspace of $\ell_2$. We also know that
our resisting oracle constructs orthonormal operators $S_k$, so that
the optimal solution of the function $f$ being constructed can be as
far away as possible from this subspace. The next proposition, which
almost establishes the theorem, essentially quantifies just how far
the optimum lies from this span.

\begin{proposition}
\label{prop:bound}
The minimum $x^*$ of function $f_{K-1}$ satisfies
\[ 
   \mathrm{dist}\inBrack{~x^*,\:\Span(S_{K-1})~} ~\geq~ \|x^*\|\:\:q^{2K} 
     \quad\text{\rm with~~} q=\frac{\sqrt{\kappa}-1}{\sqrt{\kappa}+1}~~
          \text{\rm and~~} \kappa=\frac{\mu}{L}~.
\]
\end{proposition}
\begin{proof*}
Any vector $x\in\Span(S_{K-1})$ is such that 
$(\bar{S}_{K-1})\t x \:\in\: R_{\dim(S_{K-1})} \:\subset\: R_{2K}$.

Meanwhile, equation~(\ref{eq:ro:value}) and
Proposition~\ref{prop:minimum} imply that $(\bar{S}_{K-1})\t x^* =
(\rho\,q^i)_{i=1}^\infty$.  Therefore 

$\displaystyle\mbox{\quad} \|x^*-x\|^2 = \|(\bar{S}_{K-1})\t
x^*-(\bar{S}_{K-1})\t x\|^2 ~\geq~ \sum_{i=2K+1}^{\infty}(\rho\,q^i -
0)^2 = q^{4K} \sum_{i=1}^{\infty} (\rho q^{i})^2 = q^{4K} \|x^*\|^2
~.$
\end{proof*}

\bigskip

Proposition~\ref{prop:bound} and Lemma~\ref{lemma:span} then directly
yield the theorem. Indeed, the theorem is trivial when $K=0$.
Consider otherwise an algorithm $\algB$ known to pick its answer
$x^\algB_K$ in $\Span(x_0\dots\,x_{K-1},\,f'(x_0)\dots\,f'(x_K)$. For
an appropriate choice of constant $\rho$, Proposition \ref{prop:bound}
constructs a function that satisfies the theorem.
Finally, for any algorithm $\algA$, lemma~\ref{lemma:span} implies 
that there is a function $f\in\calSzg$ such that 
\mbox{$\|x^*_f-x^\algA_K\|\geq\|x^*_f-x^\algB_K\|$}~.

Lemma~\ref{lemma:magicbound} then yields the corollary.
\begin{corollary}
\label{corr:sbound}
In order to guarantee that $\|x^*-x_K\|\leq\epsilon\|x^*\|$ for $\epsilon<1$, 
any first order black box algorithm for the optimization of $f\in\calSz$ 
must perform at least $K=\Omega(\sqrt{\kappa-1}\:\log(1/\epsilon))$
calls to the oracle.
\end{corollary}

\bigskip
Since this lower bound is established in the case where $\domain=\ell_2$, it
should be interpreted as the best \emph{dimension independent} guarantee that
can be offered by a first order black box algorithm for the optimization of
$L$-smooth $\mu$-strongly convex functions.

\end{document}